%% file: main.tex
\documentclass[10pt, conference, compsocconf]{IEEEtran}
\usepackage[utf8]{inputenc}

\ifCLASSINFOpdf
\else
\fi
\usepackage{color,soul}
\usepackage{subcaption}

\input{preamble}

\makeatletter
\let\oldtheequation\theequation
\renewcommand\tagform@[1]{\maketag@@@{\ignorespaces#1\unskip\@@italiccorr}}
\renewcommand\theequation{(\oldtheequation)}
\makeatother

\newcommand{\ie}{i.e., }

\usepackage{xcolor}
\usepackage[normalem]{ulem}
\definecolor{tangelo}{rgb}{0.98, 0.3, 0.0}

\pdfoutput=1
\begin{document}
\title{Evaluation of Skid-Steering Kinematic Models for Subarctic Environments}
\author{\IEEEauthorblockN{Dominic Baril, Vincent Grondin, Simon-Pierre Deschênes, Johann Laconte, Maxime Vaidis, \\ 
Vladim\'{i}r Kubelka, André Gallant, Philippe Giguère, François Pomerleau}
\IEEEauthorblockA{Northern Robotics Laboratory, Université Laval, Québec City, Québec, Canada\\
$\{$dominic.baril, francois.pomerleau$\}$@norlab.ulaval.ca}
}

\linepenalty=3000
\addtolength{\textfloatsep}{-0.2in}
\maketitle
\thispagestyle{plain}
\pagestyle{plain}

\begin{abstract}
In subarctic and arctic areas, large and heavy skid-steered robots are preferred for their robustness and ability to operate on difficult terrain. 
State estimation, motion control and path planning for these robots rely on accurate odometry models based on wheel velocities.
However, the state-of-the-art odometry models for \acp{SSMR} have usually been tested on relatively lightweight platforms. 
In this paper, we focus on how these models perform when deployed on a large and heavy (\SI[detect-weight=true]{590}{\kg}) \ac{SSMR}. 
We collected more than \SI[detect-weight=true]{2}{\km} of data on both snow and concrete.
We compare the ideal differential-drive, extended differential-drive, radius-of-curvature-based, and full linear kinematic models commonly deployed for \acp{SSMR}.
Each of the models is fine-tuned by searching their optimal parameters on both snow and concrete.
We then discuss the relationship between the parameters, the model tuning, and the final accuracy of the models.

\end{abstract}


\IEEEpeerreviewmaketitle

\section{Introduction}
Locomotion models are essential for many functionalities in mobile robots. %
For instance, the prediction step in Bayesian filters used in the state estimation for localization purposes heavily relies on such models.
They are also a crucial component for path planners, which use them to identify the action sequence the vehicle needs to execute to reach a specific goal state. 
More importantly for our work, locomotion models can be employed to improve the accuracy of motion controllers.
For instance, knowledge of the locomotion model enables high-speed path following for \acp{SSMR}~\citep{Huskic2019} and the use of model-predictive control algorithms~\citep{Williams2016}.
Consequently, having access to a precise and robust locomotion model is a key component of autonomy and safety in mobile robotics.

\begin{figure}[tbp]
\centering
\includegraphics[width=\linewidth]{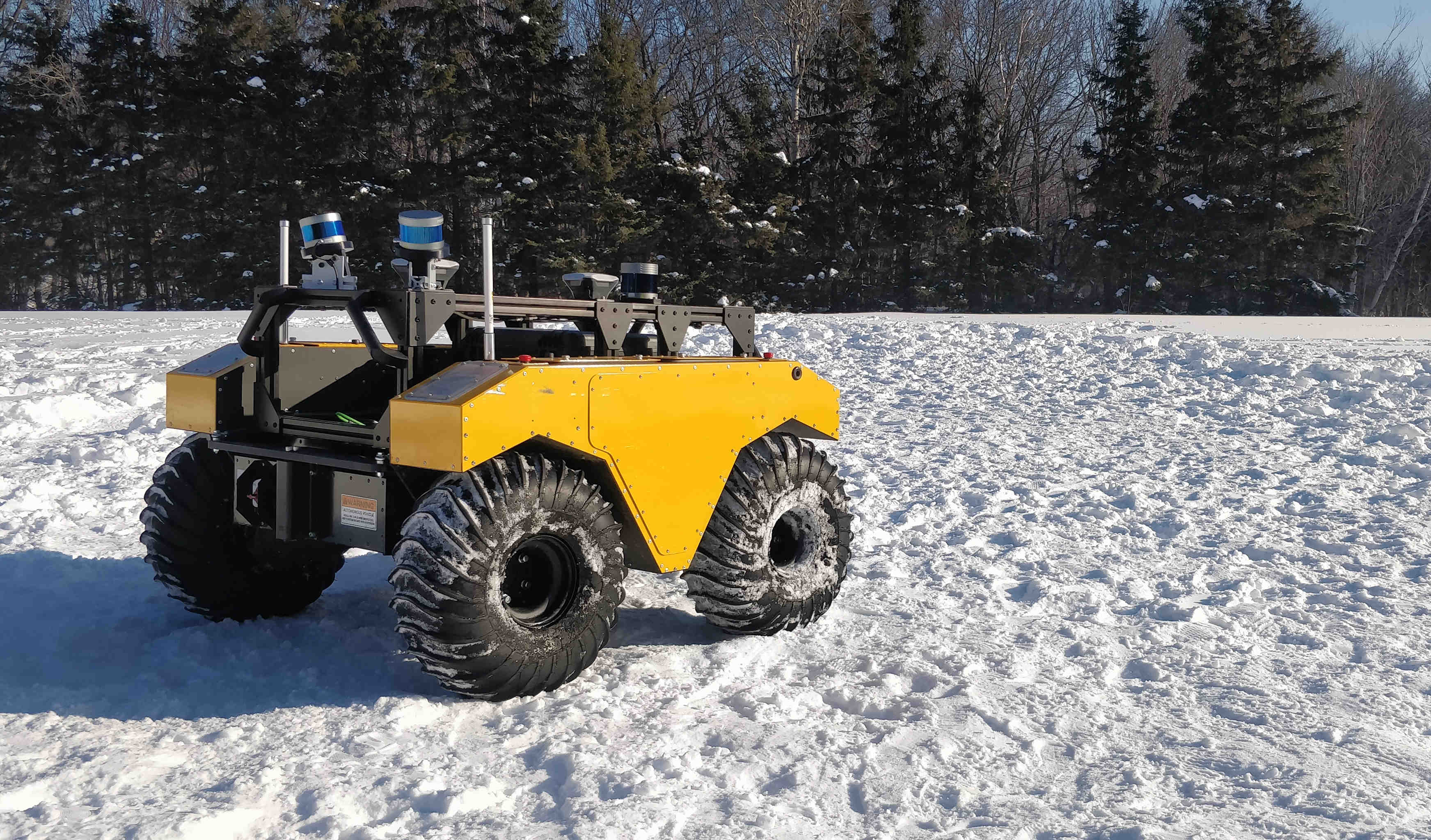}
\caption{
The wheeled skid-steer platform, a \emph{Warthog} from \emph{Clearpath}, on a snow-covered testing site. 
Due to the steering type and the snow on the ground, the robot is very prone to skidding and slippage, making the control and state estimation of this system quite challenging.
}
\label{fig:intro_picture}
\end{figure}

Different steering methods have been developed for a number of wheel geometries, a popular example being Ackerman~\citep{Shamah2001}. 
However, these wheel geometries have inherent kinematic constraints, such as a minimum turn radius.
Alternatively, the \ac{SSMR} locomotion type is designed specifically to alleviate these constraints and offers high maneuverability including zero-radius turning.
The \ac{SSMR} model operates with a set of wheels or tracks on each side of the robot, typically mechanically linked such that they have the same rotational velocity. 
The difference in velocities between the left and right wheels translates into a rotational motion of the body of the robot, much like a differential-drive system. 
An example of a wheeled \ac{SSMR}, used in this work, is shown in~\autoref{fig:intro_picture}. 
Wheeled skid-steering locomotion systems have proven to be suitable for driving at higher speeds on varying terrain~\citep{Huskic2019}.
The simplicity and robustness of the mechanical design, which includes no additional steering system, make them relatively cheap and dependable for outdoor deployments~\citep{Huskic2019}.

Kinematic models, which define the relationship between the wheel velocities and the robot velocity, are popular in the literature due to their simplicity and robustness to inaccurate parameter estimates~\cite{Rabiee2019}.
However, the inherent slippage and skidding of \acp{SSMR} render the motion difficult to predict accurately through modeling. 

In this work, we propose an experimental comparison of kinematic models applied to a heavy \ac{SSMR}. 
A special emphasis has been placed on operating on snow-covered terrain.
Because of several factors, such as uneven and unpredictable ground interaction forces, \ac{SSMR} motion on snow-covered terrain can be particularly difficult to model.
Modeling in these snowy conditions has also been little explored.
Moreover, heavy platforms, such as ours at \SI{590}{\kg}, may perform differently than lighter platforms, even on uniform terrains such as pavement or concrete.  
Indeed, to the best of our knowledge, this is the first work to study the kinematic modeling of \acp{SSMR} above \SI{120}{\kg}.

In short, the main contribution of this work is an experimental investigation of five kinematic models on challenging terrains in order to:
\begin{enumerate}
    \item validate their fitness for a heavier platform on a relatively uniform concrete terrain;
    \item evaluate their performance for snow-covered terrain using more than \SI{2}{\km} of trajectories traveled; and
    \item highlight the impact of angular motion on the accuracy of \acp{SSMR} kinematic modeling.
\end{enumerate}

\section{Related Work}
As the motion of \acp{SSMR} has been heavily studied in the literature, various kinematic models have been proven accurate to describe their motion.
However, the validation of these models has been made with rather light-weight robots, while larger and heavier robots make a more suitable choice for deployments in adverse conditions, such as snow-covered terrain because of their typically greater payload.
The question of whether these models can or cannot be transferred to such heavy robots remains open.

\label{sec:snow-deployment}
To our knowledge, relatively few robotic deployments have been conducted in snowy environments.
\citet{Apostolopoulos2000} deployed the Nomad rover, a \SI{725}{\kg} \ac{4WD} gasoline-powered vehicle, which achieved the first autonomous discoveries of Antarctic meteorites.
\citet{Ray2007} deployed the Cool Robot, a \SI{61}{\kg} solar-powered robot that drove over \SI{500}{\km} in Antarctica. 
\citet{Gifford2009} deployed MARVIN I and MARVIN II, \SI{720}{\kg} diesel-powered tracked rovers, which were used to conduct seismic and radar remote sensing of ice sheets in polar regions.
\citet{Lever2013} deployed the Yeti, an \SI{81}{\kg} \ac{4WD} electric-powered rover in Antarctica and Greenland. The platform was used to conduct autonomous \ac{GPR} surveys in polar regions.
\citet{Paton2017} used a Clearpath Robotics Grizzly \ac{UGV} (\SI{660}{\kg}) to perform autonomous route-following in unstructured and outdoor environments. They demonstrated the robustness of the algorithms through extensive field deployments spanning over \SI{26}{\km}. However, autonomous route-following in deep snow provided unsatisfactory results.

It can be seen that most deployments on snow-covered terrains were performed with relatively heavy robots. 
None of the aforementioned work on autonomous rover deployment in snow have extensively studied \acp{SSMR} motion modeling in snow-covered terrain.

Because of the inherent slippage and skidding of \acp{SSMR}, straightforward models that assume pure rolling and no slippage are not accurate enough to describe their motion~\citep{Mandow2007}. 
\citet{Mandow2007} thus proposed an extension of the ideal differential-drive model, in which each set of wheels rotate around their own \ac{ICR}. Importantly, both \ac{ICR}s are assumed to be constant for a given terrain. They also included additional parameters to take slippage into account.
All parameters for this model are identified offline empirically.
They validated their model with the Pioneer 3-AT robot, a \SI{23.6}{\kg} \ac{4WD} skid-steer platform, on asphalt using three different sets of tires.

To produce online estimates of the wheel and \acp{SSMR} \acp{ICR}, \citet{Pentzer2014} tracked them individually using an \ac{EKF}, through the inclusion of position and heading measurements. They validated their algorithm on a \SI{118}{\kg} skid-steer robot. 
\citet{Moosavian2008} and~\citet{Wang2015} proposed an experimentally-derived relationship between the radii of curvature and the amount of slippage for \acp{SSMR} motion. \citet{Wang2015} validated their approach on a wheeled \ac{SSMR}, which is a Pioneer 3-AT.

Alternatively, \citet{Anousaki2004} have proposed a general linear model. 
This model does not take into account any physical parameters of the robot and all parameters are identified offline empirically.
This model was validated on a Pioneer 2-AT robot, weighing \SI{23.6}{\kg}.

\citet{Rabiee2019} proposed a physically interpretable friction-based kinematic model, which accounts for slippage and skidding at the wheel level. 
This approach uses parameters of a dynamic friction model which are identified empirically and offline. 
The authors used a Clearpath Robotics Jackal platform (\SI{16}{\kg}) for experimental validation.

As can be seen, all of the aforementioned models are tested on relatively light platforms. However, many winter field deployments of mobile robots, such as the ones mentioned above, use heavier platforms.
Indeed, these larger and more powerful platforms are generally employed to allow for heavier payloads.

Thus, this work aims to experimentally validate the motion prediction accuracy of five kinematic models on snowy terrain, with a heavy skid-steer mobile robot.

\section{Kinematic modeling of skid-steer motion}
\begin{figure}[htbp]
\centering
\includegraphics[width=\linewidth]{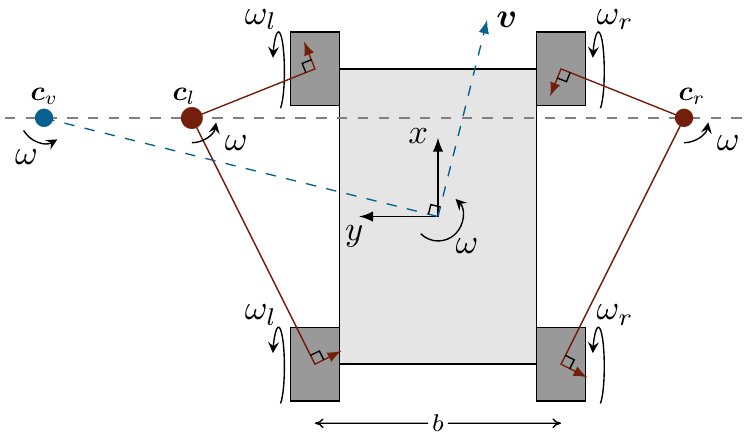}
\caption{
Diagram of a skid-steer vehicle. The instantaneous centers of rotation of the body (in blue) and tracks (in red) are shown as $\bm{c}_v$, $\bm{c}_l$ and $\bm{c}_r$.
}
\label{fig:skidsteer_diagram}
\end{figure}

Kinematic models for \acp{SSMR} aim to describe the speed of the vehicle's local frame by using two inputs: the angular velocity of the left wheel $\omega_l$ and of the right wheel $\omega_r$. 
Kinematic models do not take into account the acceleration.
Direct kinematics for the vehicle $\dot{\bm{x}}$ on the plane (i.e., in 2D) can be stated as follows:
\begin{equation}
\dot{\bm{x}} =
 \begin{bmatrix}
    \bm{v} \\
    \omega
   \end{bmatrix} =
 \begin{bmatrix}
    v_x \\
    v_y \\
    \omega
   \end{bmatrix}
 = 
  j(\omega_l, \omega_r) \\ ,
 \label{eq:diff-drive}
\end{equation}
where \(j(\cdot)\) is the kinematic model linking the inputs to the vehicle's translational velocity $\bm{v}$ and angular velocity $\omega$, as shown in \autoref{fig:skidsteer_diagram}.
The estimation of the kinematic states $\hat{\dot{\bm{x}}}$ can be computed from sensor measurements $\bm{y}$ and a Jacobian $\bm{J}$ expressed as a function of fixed parameters $\bm{k}$, such that
\begin{equation}
\hat{\dot{\bm{x}}} = \bm{J}(\bm{k}) \bm{y},
\end{equation}
with $\bm{y} = [\omega_l, \omega_r]^T$.
Based on this relation, we will define different models only by expanding the matrix $\bm{J}$ and its associated vector of parameters $\bm{k}$.

\subsection{Ideal differential-drive}
The simplest model that could be used to predict the motion of an \ac{SSMR} is an ideal differential-drive model expressed as
\renewcommand\arraystretch{1.2}
\begin{equation}
\bm{J} = r
  \begin{bmatrix}
    \phantom{-}1/2   &  1/2 \\
    \phantom{-} 0   &   0 \\
    -1/b   &   1/b  
   \end{bmatrix} ,
 \label{eq:diff-drive}
\end{equation}
where \(r\) is the radius of the wheels and $b$ is the width of the vehicle, as shown in~\autoref{fig:skidsteer_diagram}.

This model works well for two-wheeled mobile robots that have an \ac{ICR} aligned with the center of each wheel.
However, this model may be inefficient at predicting skid-steering behavior because it assumes that there is no lateral skidding or longitudinal slipping.
Nevertheless, it is easy to implement and only depends on the measurable properties of the robot such as the wheel radii and the width of the robot.

\subsection{Extended differential-drive}
The lack of skid modeling from the ideal differential-drive model pushed \citet{Mandow2007} to propose the extended differential-drive model for \acp{SSMR}. 
This model states that each set of wheels has a separate \ac{ICR}, that lies on a line parallel to the local $y$-axis also containing the \ac{ICR} of the vehicle body. 
This is shown via the red lines in \autoref{fig:skidsteer_diagram}.
This model assumes that the position of each \ac{ICR} is terrain-dependant, but constant if the terrain remains constant.
The two sets of wheels are also assumed to have the same angular velocity $\omega$ as the vehicle's body in the horizontal plane.
If one has access to the true state $\dot{\bm{x}}$, this allows the relation between the \ac{ICR}s positions and the translational and rotational velocities to be geometrically identified, as expressed by
\begin{align}
\bm{c}_v(\dot{\bm{x}}) &= 
  \begin{bmatrix}
    x_v\\
    y_v 
   \end{bmatrix} 
   = \frac{1}{\omega}
     \begin{bmatrix}
    -v_y\\
    \phantom{-}v_x 
   \end{bmatrix} \label{eq:y_icr_v}
   \\
\bm{c}_l(\dot{\bm{x}}, \omega_l)  &= 
  \begin{bmatrix}
    x_v\\
    y_l 
   \end{bmatrix} 
   = \frac{1}{\omega}
     \begin{bmatrix}
    -v_y\\
    \alpha_l (r \omega_l - v_x)
   \end{bmatrix} 
   \\  
\bm{c}_r(\dot{\bm{x}}, \omega_r)  &= 
  \begin{bmatrix}
    x_v\\
    y_r 
   \end{bmatrix} 
   = \frac{1}{\omega}
     \begin{bmatrix}
    -v_y\\
    \alpha_r (r \omega_r - v_x)
   \end{bmatrix}, \label{eq:x_icr}
\end{align}
where $\alpha_l$, $\alpha_r \in [0,1]$ are slip parameters to take into account the mechanical characteristics of the wheels~\cite{Mandow2007}.
Since we do not have access to $\dot{\bm{x}}$, as we aim at estimating it, we can use \ref{eq:y_icr_v}-\ref{eq:x_icr} to express the Jacobian of the extended differential-drive kinematic model in terms of \acp{ICR} coordinates, such that
\begin{multline}
 \bm{J}(\alpha_r, \alpha_l, x_v, y_r, y_l)
 = \\
  \frac{r}{y_{l} - y_{r}}
  \begin{bmatrix}
    -y_{r}   &  y_{l} \\
     x_{v}   &  - x_{v} \\
    -1   &   1
   \end{bmatrix} 
    \begin{bmatrix}
    \alpha_{l}   & 0 \\
    0   &  \alpha_{r} 
   \end{bmatrix} 
   .
 \label{eq:ext-diff-drive}
\end{multline}
For a symmetric robot, we can simplify the model by making the assumptions that the \acp{ICR} are symmetric concerning the center of the robot (i.e., \(y_0 = y_{l} = -y_{r}\) and \(x_{v} = 0\)) and that each set of wheels have the same slip parameter (i.e., $\alpha = \alpha_l = \alpha_r$).
This symmetric extended differential-drive model will have a Jacobian in the form of
\begin{align}
 \bm{J}(\alpha, y_0)
& =
 \frac{r\alpha}{2y_0} 
  \begin{bmatrix}
    y_0   &   y_0 \\
    0   &   0 \\
    -1   & 1
   \end{bmatrix}
   =
  r\alpha 
  \begin{bmatrix}
    1/2  &  1/2 \\
    0   &   0 \\
    -1/\hat{b}   & 1/\hat{b}
   \end{bmatrix}
 \label{eq:ext-diff-drive-sym}
   \\
    &= \bm{J}(\alpha, \hat{b}) \quad \text{with } \hat{b}= 2y_0 
    \label{eq:ext-diff-drive-sym-alt}.
\end{align}
In this case, the model only has two parameters to train, a slip parameter $\alpha$ and an estimated virtual width of the vehicle $\hat{b}$.
As with the ideal differential-drive model, the symmetric extended differential-drive model from \autoref{eq:ext-diff-drive-sym-alt} still assumes that there is no lateral skidding (i.e., \(v_y = 0\) ), but it is capable of modeling longitudinal slipping and loss of energy while steering.
In this work, both the five-parameter and the symmetric two-parameter extended differential drive models are examined.

\subsection{ROC-based}
\citet{Wang2015} experimented with \ac{SSMR} to find a relation between slippage and the \ac{ROC} of the motion. Looking at the previous model from \autoref{eq:ext-diff-drive-sym}, since \(v_y = 0\), it can be seen that the instantaneous radius of curvature of the robot is %
\begin{equation}
    R = \frac{v_x}{\omega} = \abs{\frac{\omega_r + \omega_l}{\omega_r - \omega_l}}y_0 = \lambda y_0, 
\end{equation}
where \(\lambda\) is the time-varying path curvature variable. 
Through experiments, \citet{Wang2015} identified the following relation between \(y_0\) and \(\lambda\) with
\begin{equation}
    y_0 = \frac{b}{2}\left(1+\frac{\beta_1}{1+\beta_2\sqrt{\lambda}}\right),
    \label{eq:roc-base}
\end{equation}
where $\beta_1$ and $\beta_2$ are parameters trained experimentally, giving the new Jacobian $ \bm{J}(\alpha, \beta_1, \beta_2)$ following \autoref{eq:ext-diff-drive-sym} and \autoref{eq:roc-base}.
Since \(\lambda\) is time-varying, the model adapts as a function of the \ac{ROC}, in contrast with the other models. 
However, this model still does not address lateral skidding.

\subsection{Full linear}
\citet{Anousaki2004} proposed a general linear model to account for some of the system's uncertainty and asymmetry inherent to \ac{SSMR} expressed as 
\begin{gather}
\bm{J}(\gamma_{11}, \cdots, \gamma_{32})
 =
 \begin{bmatrix}
    \gamma_{11} & \gamma_{12}\\
    \gamma_{21} & \gamma_{22}\\
    \gamma_{31} & \gamma_{32}
   \end{bmatrix},
 \label{eq:full-linear}
\end{gather}
where $\gamma_{ij}$ are the linear coefficients to be trained, leading to a model with six parameters. %
Unlike the previous models, this model requires no \textit{a priori} knowledge of the system, %
and relies entirely on parameters estimated through training. %

\section{Experimental setup}
\label{sec:exp_setup}
    To compare the presented models, we collected data using a Clearpath Warthog \ac{UGV}, as shown in~\autoref{fig:intro_picture}.
    Weighing \SI{590}{\kg}, its dimensions are \SI{1.52 x 1.38 x 0.83}{\m} and can reach a top speed of \SI[per-mode=symbol]{18}{\km\per\hour}.
    As the robot has a skid-steering locomotion system, the wheels on each side of the robot are mechanically linked to a single motor.
    The robot is also equipped with a differential suspension, stabilizing the sensors and improving wheel-ground contact. 
    A Robosense RS-32 lidar located at the front of the platform is used for localization.
    During the experiments, the lidar produced point clouds at \SI{10}{\hertz}, the wheel velocity commands are sent at \SI{20}{\hertz} and the \ac{IMU} returns readings at \SI{400}{\hertz}.
    The recorded data is then used for generating the ground truth trajectory using the \ac{ICP} algorithm.
    Importantly, a different trajectory was recorded for model training and validation.

\begin{figure}[htbp]
    \centering
        \begin{subfigure}[t]{0.23\textwidth}
        \centering
        \includegraphics[width=\linewidth]{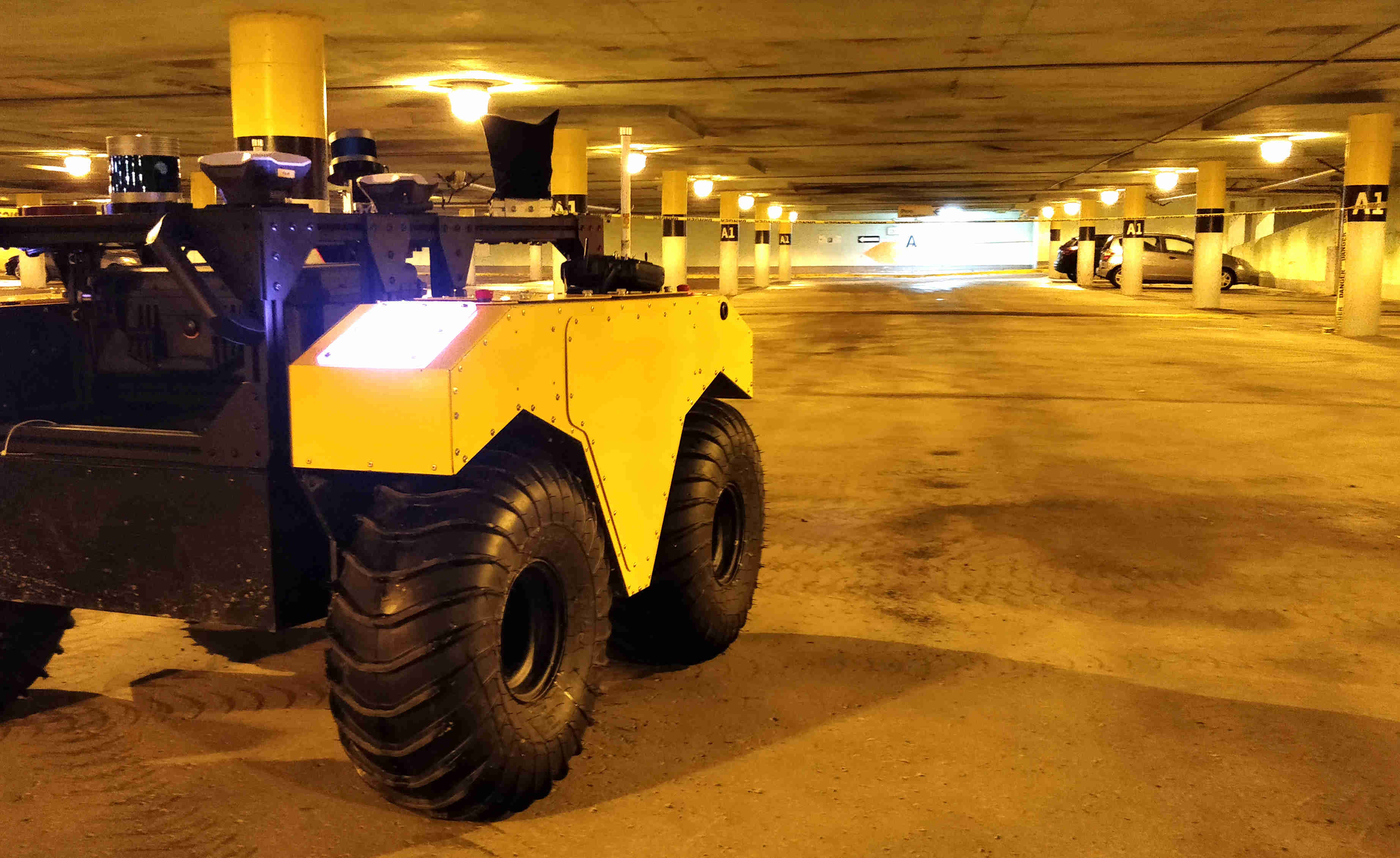}
        \caption{}
        \label{fig:env_asphalt}
    \end{subfigure}
    ~ 
    \begin{subfigure}[t]{0.23\textwidth}
        \centering
    \includegraphics[width=\linewidth]{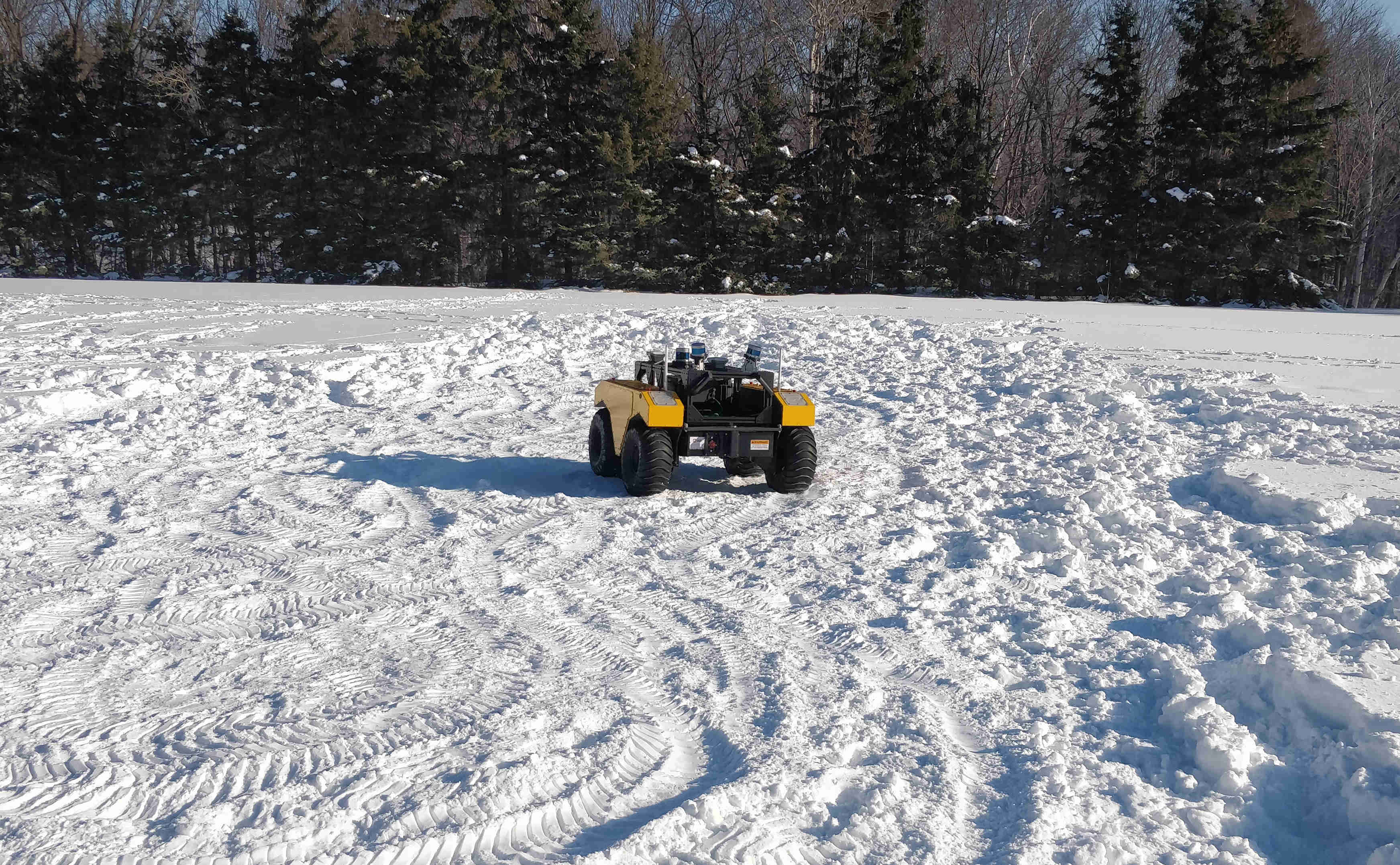}
        \caption{}
        \label{fig:env_snow}
    \end{subfigure}
    ~
    \begin{subfigure}[t]{0.23\textwidth}
        \centering
    \includegraphics[width=\linewidth, trim=300 450 200 200, clip]{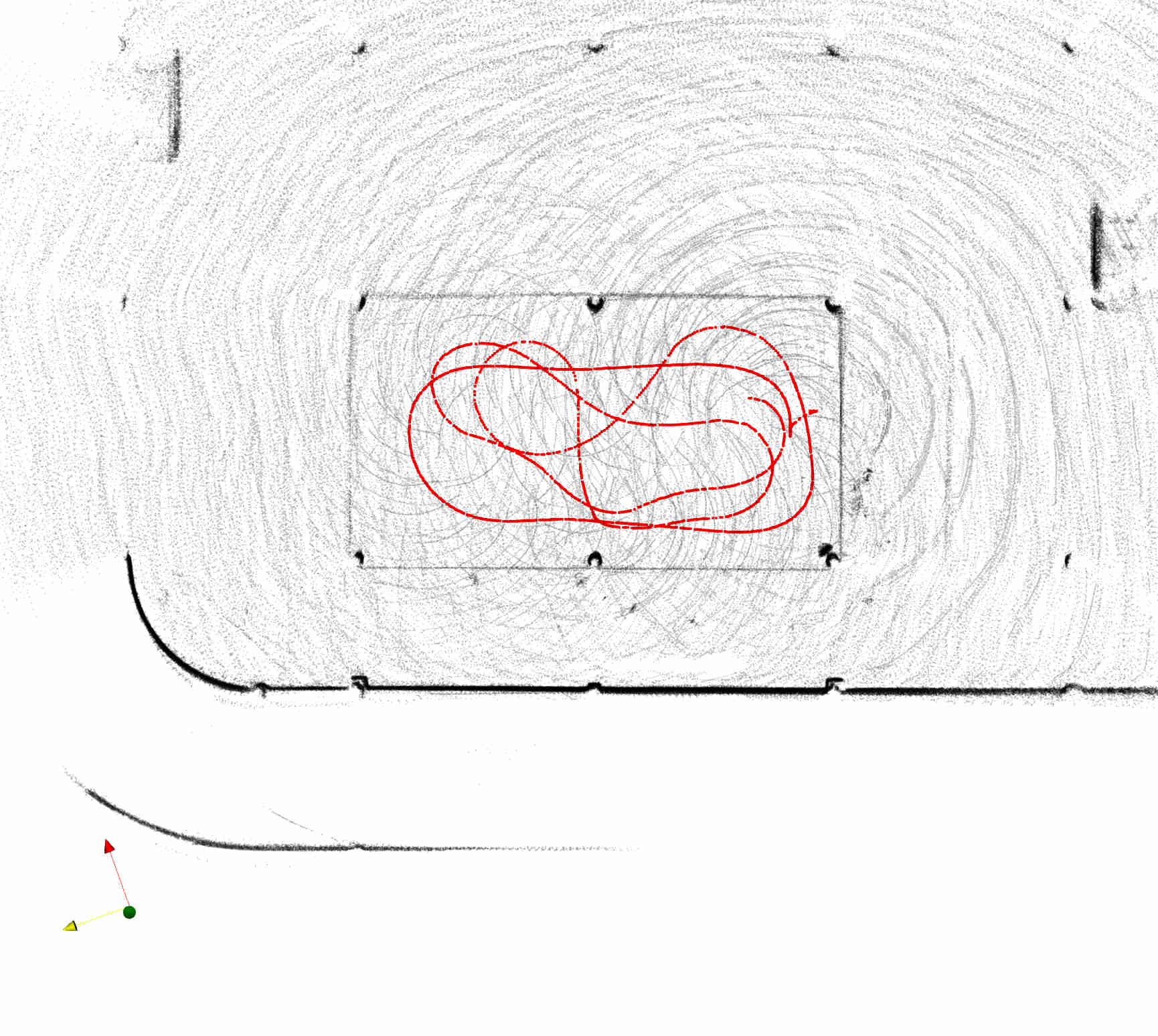}
        \caption{}
        \label{fig:icp_asphalt}
    \end{subfigure}
    ~
    \begin{subfigure}[t]{0.23\textwidth}
        \centering
    \includegraphics[width=\linewidth, trim=300 400 300 300, clip]{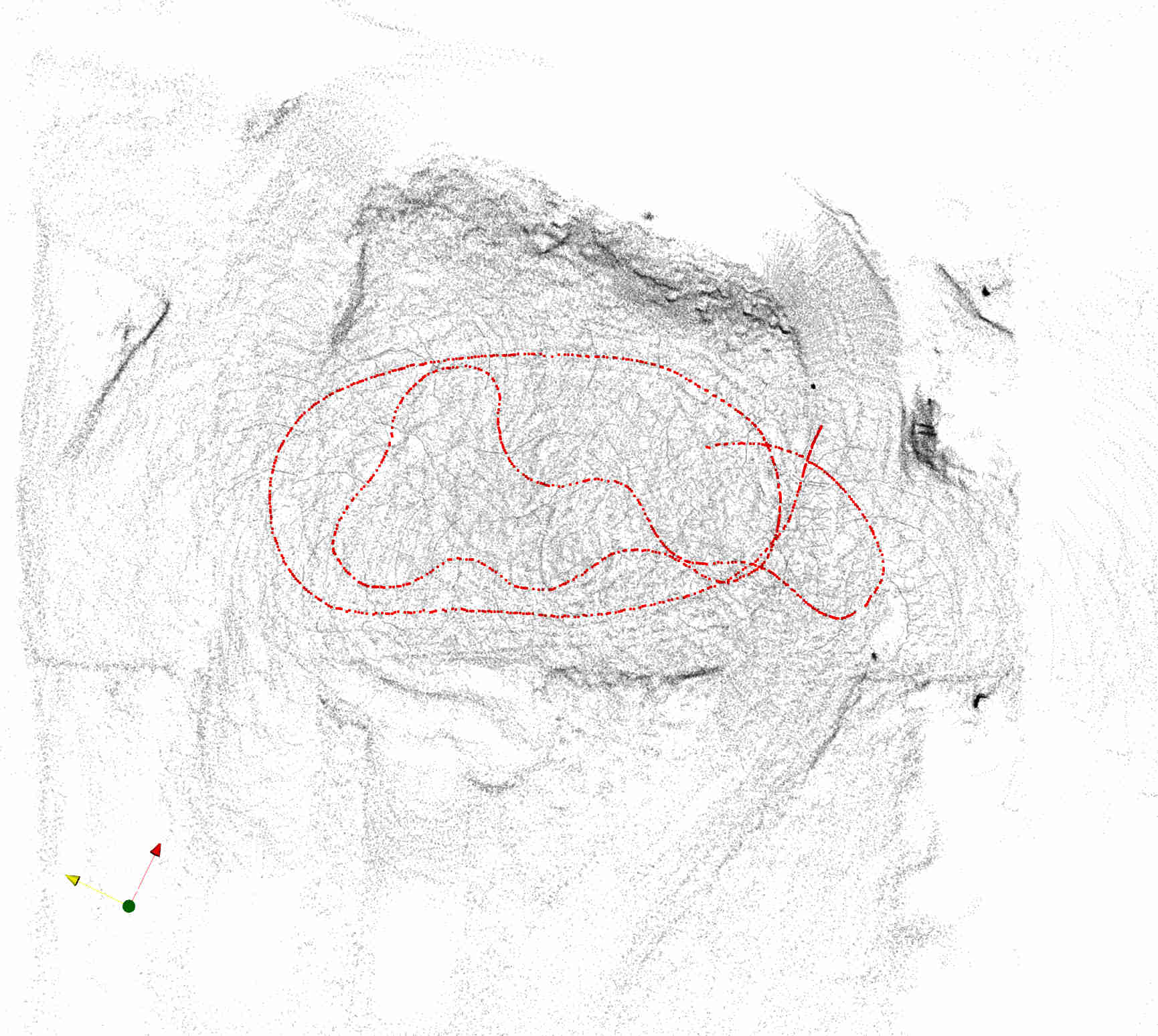}
        \caption{}
        \label{fig:icp_snow}
    \end{subfigure}
    \caption{Different road surface conditions used in experiments.
    (a) Underground parking lot with a dry concrete surface. 
    (b) Snow-covered terrain. 
    (c) The resulting map of the underground parking lot with one of the robot's trajectories plotted in red. The experimental area is surrounded by concrete pillars connected by a safety tape.    
    (d) The resulting map of the snow-covered terrain with one of the robot's trajectories plotted in red.}
    \label{fig:environments}
\end{figure}

    We tested the models on two different types of surface: a flat concrete surface and a snow-covered terrain.
    As can be seen in \autoref{fig:env_asphalt} and \autoref{fig:env_snow}, these environments present radically different physical properties.
    In the first one, the robot drives on flat and dry concrete in an underground parking lot. 
    Due to the high friction coefficient and the hardness of the ground, the skid-steering motion is induced by wheel deformation.
    Furthermore, the stick-slip phenomenon introduces additional unmodeled noise.

    On the other hand, the snow-covered terrain is soft and exhibits low friction, meaning that the skid-steering motion is induced by terrain deformation.
    Additionally, %
    the terrain unevenness adds unmodeled noise to the skid-steering motion when operating at high speeds. 
    To better train and evaluate all of the the models, the trajectories were planned in a way to maximize the excitement range and coverage of the model input variables (\ie left and right wheels commanded angular velocities).
    Some parts of these trajectories obtained by the \ac{ICP} mapping algorithm are presented in \autoref{fig:icp_asphalt} and \autoref{fig:icp_snow}.

In order to obtain an accurate location of the platform to compare the different models, we used the \ac{ICP} algorithm.
This algorithm allows for accurate odometry measurements, by registering 3D point clouds together~\citep{Pomerleau2013}.
Because of its few centimeters margin of error, it was not necessary to resort to more precise measuring tools such as theodolites. 
Moreover, since each platform trajectory is estimated with identical \ac{ICP} parameters, the magnitude of the localization error will be the same for each of the models tested. 
As a result, the \ac{ICP} positioning and orientation errors will not change the behavior of the different models tested.
The library \texttt{libpointmatcher}~\citep{Pomerleau2014} was used to compute \ac{ICP} offline, using a point-to-plane minimization.

To properly evaluate each model, it is necessary to train for model parameters and evaluate model performance on two separate trajectories.
Our model training procedure is similar to the two-step method proposed in \citep{Mandow2007}.
The first step is to obtain experimental data by driving the robot manually, while recording commanded wheel velocities and sensor measurements.
The data is then processed offline to compute the ground truth localization, using the \ac{ICP} algorithm. The training path is then split into $N$ distinct segments corresponding to a total traveled distance of $h_t$. 
The commanded wheel velocities corresponding to each of the $N$ segments are then used to predict the robot's motion, starting from the initial position of the corresponding segment.
The final model-predicted position of the robot is then compared with the ground truth position for each segment.
Our loss function $l(\cdot)$ is the sum of $N$ squared Mahalanobis distances using
\begin{equation}
    l(\bm{k}) = \sum_{i=1}^{N} (\bm{x} - \hat{\bm{x}})^T \bm{\Sigma}^{-1} (\bm{x} - \hat{\bm{x}}) ,
    \label{eq:objective_fct}
\end{equation} %
where $\bm{x}$ and $\hat{\bm{x}}$ are respectively the ground truth and model-predicted state of the robot defined as its 2D position and its orientation.
The covariance matrix $\bm{\Sigma}$ is there to bring meters for the position and radians for the orientation in a common unitless value.
In our case, we set this covariance to identity.
The set of parameters for each model is represented by the vector $\bm{k}$.
An optimization algorithm is then used to search for the set of parameters $\bm{k}$ minimizing the loss function $l(\cdot)$. 

Previous work on model parameter identification for \ac{SSMR} relied on temporal horizons for ground truth trajectory segmentation \citep{Mandow2007, Rabiee2019, Seegmiller2014}.
We investigated two different strategies when splitting the ground truth trajectory: by using a temporal and a spatial horizon.
In particular, we found that using a spatial horizon for model parameter optimization allowed for the easy removal of outlier data generated when zero velocity commands were given to the robot.
The trained models are then evaluated using two different metrics: the relative error of the translational prediction $\varepsilon_t$ and the relative error of the angular prediction $\varepsilon_\theta$, respectively computed as the translational error and angular error the predictor does per meter.
These two metrics are computed on the evaluation horizon $h_e$ over the entire evaluation trajectories, %
as in \citep{Geiger2012}.

\section{Results}
In order to fine-tune and evaluate the different models, we first determine the impact of the training and evaluation horizon $h_t$ and $h_e$ on the model performance. 
Once the best horizons are chosen, the models are trained and compared in both linear and angular errors.
We then deepen our study with the most promising model, the differential drive symmetric model, by looking at the relation between the errors and the commands sent to the robot.

\begin{figure}[htbp]
    \centering
    \includegraphics[]{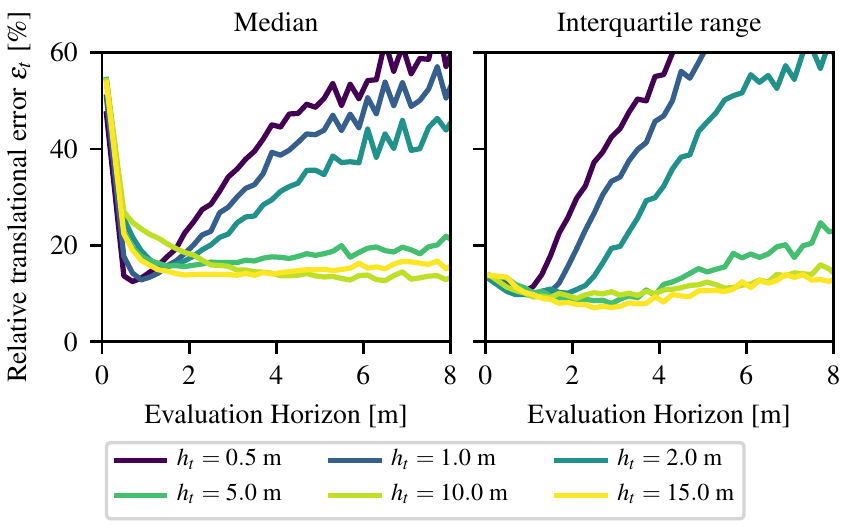}
    \caption{
    Relative translational error $\varepsilon_t$ of the extended differential-drive asymmetric model in a snowy environment, as a function of the evaluation horizon window $h_e$ for different training windows $h_t$.
    \textit{Left:} Median of the relative linear error. \textit{Right:} Interquatile range of the relative linear error.
    }
    \label{fig:horizon_error_snow2}
    \end{figure}

In order to determine the impact of the training horizon $h_t$ and the evaluation horizon $h_e$ on the performance, we trained every model for various values for $h_t$ and evaluated the translational error $\varepsilon_t$ for various horizons $h_e$. Except for the \ac{ROC}-based model, which did not react to a variation in training horizon $h_t$, the error $\varepsilon_t$ of the different models had a similar behavior when the values for the horizons $h_t$ and $h_e$ were varying.
As an example, the translational error $\varepsilon_t$ values of the extended differential-drive asymmetric model prediction for varying training and evaluation horizons are shown in~\autoref{fig:horizon_error_snow2}.

We can first observe that the model error $\varepsilon_t$ behaves differently depending on the training horizon $h_t$.
Models trained on smaller horizons offer better performances at lower evaluation horizons while they suffer at higher evaluation horizons.
This indicates that the model training horizon $h_t$ should be chosen in accordance with the application of the model.
In robotics applications, controllers mainly use small horizon.
Thus, the training and the evaluation horizons should mainly use small horizons.

It can be seen in~\autoref{fig:horizon_error_snow2} that training horizons of \SI{2}{\m} or \SI{5}{\m} quickly converge to a small error compared to longer training horizons, while not suffering of high error for long evaluation horizons as much as smaller training horizons.
For this model, the \SI{15}{\m} $h_t$ also shows quick convergence but this effect is only observed for the  extended  differential-drive  asymmetric  model on snow.
Furthermore, all the median curves assume similar values at $h_e=\SI{2}{\m}$.
The initial error for $h_e$ approaching $\SI{0}{\m}$ is caused by the combined measurement noise of the wheel velocities and our ground truth. 
Hence, taking an evaluation horizon of \SI{2}{\m} leads to a certain invariance of the error $\varepsilon_t$ on the training horizons. 
Also, longer evaluation horizons $h_e$ tend to introduce a high, interquartile range for predicting the errors $\varepsilon_t$, meaning that the evaluation is very dependant on the state of the robot.
As we want to evaluate the models on complex trajectories, an evaluation horizon of $h_e=\SI{2}{\m}$ is chosen for the rest of this work.
The evaluation trajectories differ from the training trajectory but meet the same goal of maximizing model input excitement range and coverage.
We have used this $h_t$ to train for all model parameters using the parameter identification method described in~\autoref{sec:exp_setup}.
The overview of the parameters can be found in \autoref{tab:param}. 
    
\begin{table}[htpb]
\centering
\caption{Summary of parameters used and trained for each model.
For all models, the wheel radius $r$ is equal to $\SI{0.3}{\m}$.}
\begin{tabu}{@{}X  l l l@{}} 
\toprule
\emph{Model} & \emph{Trained concrete} & \emph{Trained snow} & \emph{Bounds} \\
\midrule
DD with $b=1.2$ m 
& -- & --  & -- \\
\midrule
Extended
& $\alpha= \phantom{-}0.94$ & $\alpha= \phantom{-}0.86$ & $\in [0, 1]$ \\
DD Symmetric& $\hat{b}= \phantom{-}\SI{4.46}{\m}$ & $\hat{b}= \phantom{-}\SI{3.08}{\m}$ & $\in [0, \infty)$ \\
\midrule
Extended
& $\alpha_l= \phantom{-}0.90$ & $\alpha_l= \phantom{-}0.81$ & $\in [0, 1]$ \\
DD Asymmetric & $\alpha_r= \phantom{-}0.92$ &  $\alpha_r= \phantom{-}0.84$ & $\in [0, 1]$ \\
& $x_v= \SI{-2.57}{\m}$ & $x_v= \SI{-2.71}{\m}$ & $\in \mathbb{R}$ \\
& $y_l= \phantom{-}\SI{4.66}{\m}$ & $y_l= \phantom{-}\SI{3.00}{\m}$ & $\in [0, \infty)$ \\
& $y_r= -\SI{5.00}{\m}$ & $y_r= -\SI{3.85}{\m}$ & $\in (-\infty, 0]$ \\
\midrule
ROC 
& $\alpha= \phantom{-}0.91$ & $\alpha_l= \phantom{-}0.80$ & $\in [0, 1]$ \\
with $b=1.2$ m  & $\beta_1= 42.73$ & $\beta_1=  \phantom{-}1.36$ & $\in \mathbb{R}$ \\
& $\beta_1= 11.09$ & $\beta_2= -0.18$ & $\in \mathbb{R}$ \\
\midrule
Full linear
& $\gamma_{11}= \phantom{-}0.47$ & $\gamma_{11}= \phantom{-}0.46$ & $\in \mathbb{R}$ \\
& $\gamma_{12}= \phantom{-}0.44$ & $\gamma_{12}= \phantom{-}0.36$ & $\in \mathbb{R}$ \\
& $\gamma_{21}= -0.22$ & $\gamma_{21}= -0.31$ & $\in \mathbb{R}$ \\
& $\gamma_{22}=\phantom{-}0.26 $ & $\gamma_{22}= \phantom{-}0.34$ & $\in \mathbb{R}$ \\
& $\gamma_{31}= -0.10$ & $\gamma_{31}= -0.13$ & $\in (-\infty, 0]$ \\
& $\gamma_{32}= \phantom{-}0.08$ & $\gamma_{32}= \phantom{-}0.12$& $\in [0, \infty)$ \\
\bottomrule
\end{tabu}
\vspace{5pt}
\footnotesize{Legend: DD = Differential drive.}
\label{tab:param}
\end{table}

\begin{figure}[htbp]
    \centering
    \includegraphics[]{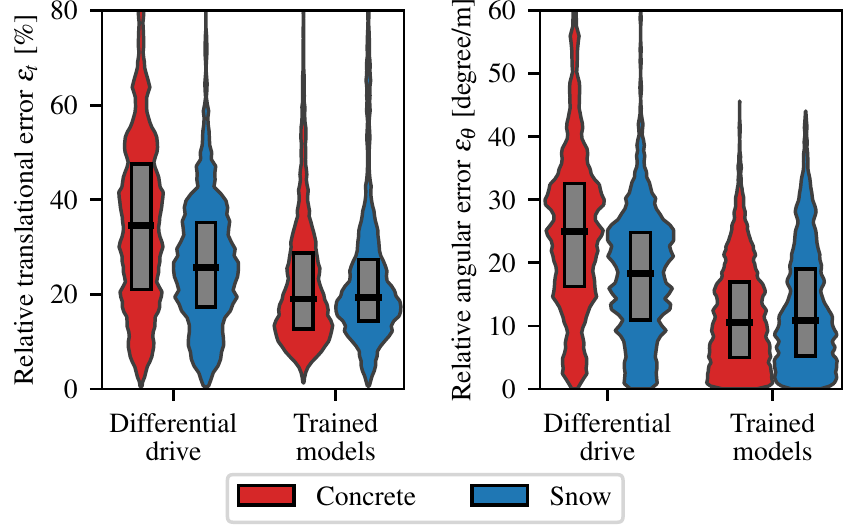}
    \caption{
    Overall errors of the differential-drive model and the trained models, combined in a single distribution. As expected, the trained models perform vastly better, as they take into account the wheel ground contact interactions.
    }
    \label{fig:overall_error_DD_all}
    \end{figure}

\begin{figure*}[htbp]
    \centering
    \includegraphics[]{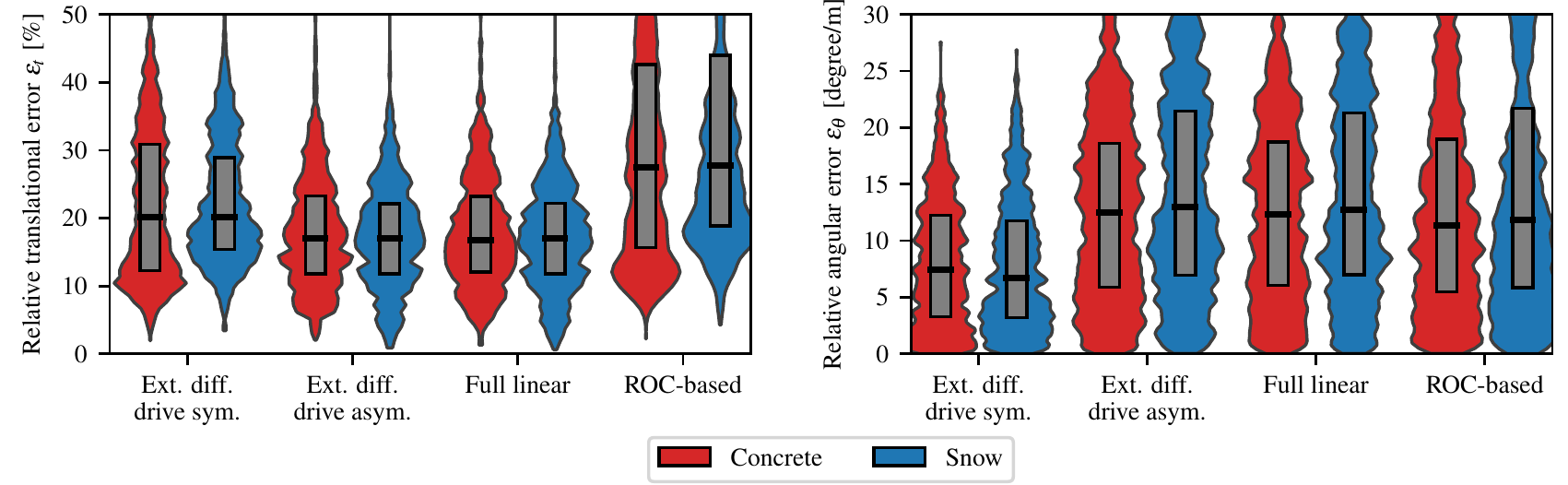}
    \caption{
    Overall error of the models depending on which environment the robot evolved in.
    }
    \label{fig:overall_error}
    \vspace{-5mm}
    \end{figure*}

    The translational errors $\varepsilon_t$ and angular errors $\varepsilon_\theta$ for this experiment are shown in~\autoref{fig:overall_error_DD_all} and \autoref{fig:overall_error}.
    In them, the median and quartiles at \SI{25}{\%} and \SI{75}{\%} are depicted with the gray box, while the underlying curves show the data distribution.
    In \autoref{fig:overall_error_DD_all}, we can see that the residual errors for the ideal differential-drive model are much higher than for any of the trained models.
    Indeed, the differential-drive model does not take the slippage and skidding phenomenom, leading to high errors for \acp{SSMR}.
    It can be observed that the ideal differential drive performs better on snow that on concrete. 
    This shows that \acp{SSMR} motion is closer to that of an ideal differentially driven robot when operating on snow-covered terrain than when operating on concrete.
    
    \autoref{fig:overall_error} shows in greater details the distribution of errors for the four trained models. 
    There, it can be seen that except for the \ac{ROC}-based one, all models tend to perform similarly. 
    The model prediction error is similar for both snow (blue) and concrete (red).
    Furthermore, the extended  differential-drive asymmetric and full linear models offer the best linear displacement predictions, which is due to the fact that they account for lateral motion.
    However, the extended differential-drive symmetric model offers more accurate angular displacement prediction while offering slightly inferior accuracy for translational displacement prediction than other kinematic models presented in this work. 
    The fact that it only has two parameters to be trained makes it an interesting choice for skid-steering locomotion modeling.
    Indeed, fewer parameters lead to a smaller computational cost for training, and such a model is less prone to converge into a local minimum as the search space is smaller.
    Furthermore, richer models can also suffer from overfitting on the training data, which is less likely to happen with fewer parameter models.

    It should be noted that the symmetry hypothesis underlying many models is very close to reality for our platform.
    This is understandable, as our vehicle is almost symmetric in design and the added components add negligible mass to the system.
    A \ac{SSMR} not respecting this constraint would require the use of a model that allows for the asymmetry, such as the extended differential-drive asymmetric model or the full linear model, but at the cost of a higher model dimensionality, therefore bearing the aforementioned disadvantages.

\begin{figure}[htbp]
    \centering
    \includegraphics[]{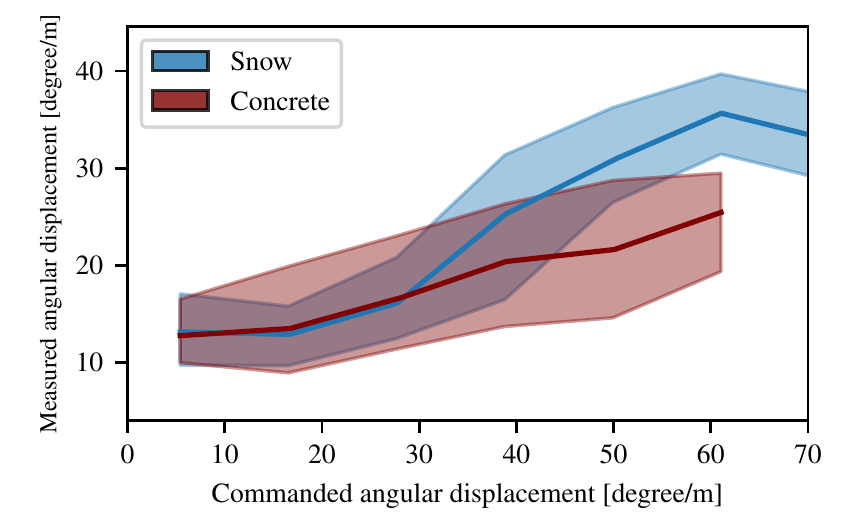}
    \caption{
    Actual rotation over one meter as a function of the difference of wheel velocity.
    The median is represented with a straight line, where the shaded zones represent the quartiles at \SI{25}{\%} and \SI{75}{\%}.
    Because of the low friction coefficient of the snow, the robot tends to rotate more for the same command compared to concrete.
    }
    \label{fig:skiddingSnow}
    \end{figure}

    In order to highlight the differences of wheel-ground interactions between the snow and the concrete, we measured the actual rotation of the robot for a series of given commands.
    \autoref{fig:skiddingSnow} shows the measured angular displacement given the commanded angular displacement for the evaluation trajectories on snow and concrete.
    As a lot of data was collected, the median and the quartiles at 25\% and 75\% were computed to ease the reading.
    
    It can be seen that for low commanded angular displacement, the robot behaves similarly on snow and on concrete.
    However, for commanded angular displacements of over \SI{30}{\degree}, the curve for resulting angular displacement grows faster when the robot is operated on snow than on concrete.
    A drop in measured angular displacement for high angular displacement commands can also be seen for snow-covered terrain.
    This result suggests a nonlinear effect in \acp{SSMR} motion on snow-covered terrain which could be hard to describe with linear kinematic models.
    Additionally, interquartile range is generally stable for both terrain types but higher on concrete than on snow, showing that for a specific angular displacement command, the range of possible angular displacement actually done by the \ac{SSMR} is higher on concrete than on snow.
    Higher speeds were not reached for the concrete trajectory because of safety purposes. 
    The lack of data for zero commanded angular velocity is due to a combination between the choice of a spatial evaluation window $h_e$ = \SI{2}{\m} and the trajectory planning aiming to maximize commanded wheel velocities range and coverage, leading to no window with zero commanded angular displacement.
    It can be observed that for small angular displacement commands, the robot can have a higher angular displacement than the commanded one.
    This could be due to two phenomena: as the robot was driven at high velocities on uneven, snow-covered terrain, the suspension was not able to compensate for the vibrations leading to a loss of contact between the wheels and the ground. 
    This additional noise in the command led the robot to slightly turn in straight lines.
    Also, the high momentum of the robot sometimes caused it to continue to turn in end of turns even if the actual command had a null angular displacement.
    In addition, the relative translational displacement is constant for every commanded linear displacement and is the same for snow and concrete.
    This shows that the angular displacement is the main factor of error while commanding \acp{SSMR}.

    As demonstrated before, the best model in our case is the extended differential drive symmetric.
    The study of this model is then deepened to the study of the angular error.
    Indeed, the linear error was independent of the given commands in our study.
    \begin{figure}[htbp]
        \centering
        \includegraphics{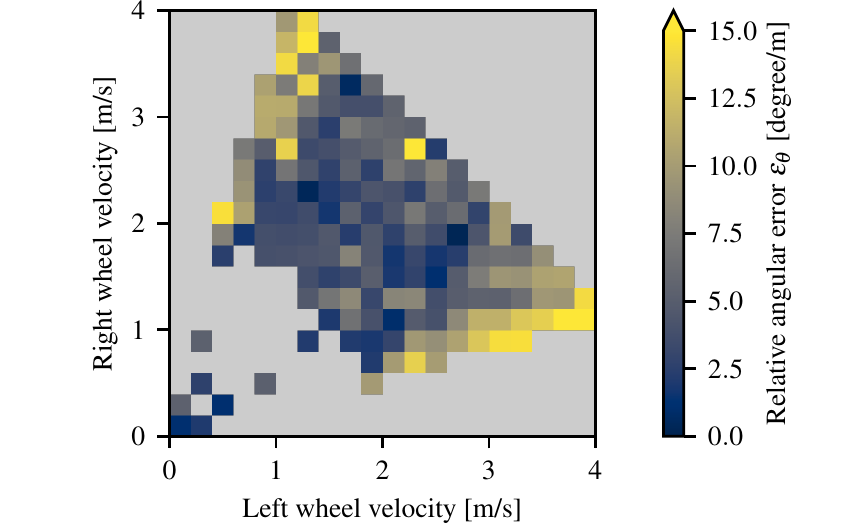} %
        \caption{Relative angular error for the extended differential drive symetric model on snow. We observed a similiar behavior on the concrete. }
        \label{fig:errorAng_eddS_snow}
    \end{figure}
    
    An analysis of relative errors in prediction was conducted for the extended differential-drive symmetric model on both snow and concrete.
    The prediction error was computed for every evaluation trajectory segments, as well as the corresponding wheels angular velocities commands.
    This way, we were able to determine that the relative translational prediction errors $\varepsilon_t$ were uncorrelated with mean angular velocity commands for each side.
    However, angular prediction errors $\varepsilon_\theta$ were correlated with mean angular velocity commands, as can be seen in~\autoref{fig:errorAng_eddS_snow}.
    It can be observed that the prediction error reaches its maximum in the areas when either side's commanded wheel velocity is about twice that of the other side.
    In particular, we see a steep increase in the prediction relative angular error in the high error areas. 
    This suggests a non-linearity in the relationship, which cannot be captured by any of the linear models tested.

\section{Conclusion}
In this paper, we compared five different kinematic models to describe the motion of a \SI{590}{\kg} \ac{SSMR} platform both on concrete and on a snow-covered terrain.
We have shown that the model training horizon should be selected based on the model's application.
We have compared the prediction accuracy of five kinematic models for \acp{SSMR}.
We have also highlighted differences in the behavior and model-prediction error of \acp{SSMR} when operating on the two different terrain types.
We have also identified commanded angular wheel velocity sets that induce high prediction errors of angular displacement.
As future work, we aim to implement more advanced models and to develop a path-following algorithm robust to multiple terrain types for a given trajectory.

\section*{Acknowledgements}

This research was supported by the Natural Sciences and Engineering Research Council of Canada (NSERC) through the grant CRDPJ 511843-17 BRITE (bus rapid transit system), CRDPJ 527642-18 SNOW (Self-driving Navigation Optimized for Winter) and FORAC.

\IEEEtriggeratref{6}
\IEEEtriggercmd{\enlargethispage{-0.1in}}

\printbibliography

\end{document}

%% file: preamble.tex
\usepackage[
backend=bibtex8, 
style=ieee, 
sorting=none, 
natbib=true, 
doi=false, 
isbn=false, 
url=false, 
eprint=false, 
maxcitenames=1, 
mincitenames=1
]{biblatex}

\usepackage[pdftex,colorlinks]{hyperref}
\usepackage[printonlyused]{acronym}
\acrodef{SSMR}{skid-steer mobile robot}
\acrodef{ICR}{instantaneous center of rotation}
\acrodef{ROC}{radius of curvature}
\acrodef{GPR}{ground-penetrating radar}
\acrodef{EKF}{extended Kalman filter}
\acrodef{4WD}{four-wheel drive}
\acrodef{ICP}{Iterative-Closest-Point}
\acrodef{UGV}{unmanned ground vehicle}
\acrodef{ROS}{Robot Operating System}
\acrodef{IMU}{Inertial Measurement Unit}
\acrodef{RMSRE}{root mean squared relative error}

\usepackage{siunitx}
\usepackage[all]{nowidow}

\usepackage[dvipsnames, table]{xcolor}

\usepackage{lipsum}

\usepackage[pdftex]{graphicx}

\usepackage{epstopdf}

\usepackage{import}

\graphicspath{{./latexGoodPractices/}}

\usepackage{booktabs}

\usepackage{tabu}

\usepackage{amssymb,amsfonts,amsmath,amscd}

\usepackage{bm}

\newcommand{\abs}[1]{\left\vert#1\right\vert}

\newcommand{\bbm}{\begin{bmatrix}}
\newcommand{\ebm}{\end{bmatrix}}